\documentclass[preprint,12pt]{elsarticle}

\usepackage{amssymb}
\usepackage{amsmath}
\usepackage{xcolor}
\usepackage{subcaption}
\usepackage{tabularx}
\usepackage{booktabs}
\usepackage{multirow}
\usepackage{colortbl}
\usepackage[colorlinks]{hyperref}
\usepackage{url}

\usepackage{lineno}

\definecolor{Gray}{gray}{0.85}
\definecolor{LightCyan}{rgb}{0.88,1,1}
\definecolor{antiquewhite}{rgb}{0.98, 0.92, 0.84}
\definecolor{atomictangerine}{rgb}{1.0, 0.6, 0.4}
\newcolumntype{Y}{>{\centering\arraybackslash}X}

\makeatletter
\def\ps@pprintTitle{%
    \let\@oddhead\@empty
    \let\@evenhead\@empty
    \def\@oddfoot{\footnotesize\itshape
         {Preprint submitted to Image and Vision Computing} \hfill}%
    \let\@evenfoot\@oddfoot
    }
\makeatother

\journal{Image and Vision Computing}

\begin{document}

\begin{frontmatter}



\title{One-shot lip-based biometric authentication: extending behavioral features with authentication phrase information}


\author[inst1]{Brando Koch}
\ead{bkoch4142@gmail.com}

\author[inst1]{Ratko Grbi\'{c}\corref{cor1}}
\ead{ratko.grbic@ferit.hr}
\cortext[cor1]{Corresponding author.}

\affiliation[inst1]{organization={Faculty of Electrical Engineering, Computer Science and Information Technology Osijek},
            addressline={Kneza Trpimira 2B}, 
            city={Osijek},
            postcode={HR-31000}, 
            country={Croatia}}

\begin{abstract}
Lip-based biometric authentication (LBBA) is an authentication method based on a person’s lip movements during speech in the form of video data captured by a camera sensor. LBBA can utilize both physical and behavioral characteristics of lip movements without requiring any additional sensory equipment apart from an RGB camera.
State-of-the-art (SOTA) approaches use one-shot learning to train deep siamese neural networks which produce an embedding vector out of these features. Embeddings are further used to compute the similarity between an enrolled user and a user being authenticated. A flaw of these approaches is that they model behavioral features as style-of-speech without relation to what is being said. This makes the system vulnerable to video replay attacks of the client speaking any phrase. 
To solve this problem we propose a one-shot approach which models behavioral features to discriminate against what is being said in addition to style-of-speech. We achieve this by customizing the GRID dataset to obtain required triplets and training a siamese neural network based on 3D convolutions and recurrent neural network layers. A custom triplet loss for batch-wise hard-negative mining is proposed. Obtained results using an open-set protocol are 3.2\% \textit{FAR} and 3.8\% \textit{FRR} on the test set of the customized GRID dataset. Additional analysis of the results was done to quantify the influence and discriminatory power of behavioral and physical features for LBBA.
 
\end{abstract}

%\includegraphics[width=0.9\linewidth]{imgs/graphical_abstract.png}

%\item One-shot approach applied to lip-based authentication
%\item Physical and behavioral feature biometrics
%\item Customized GRID dataset
%\item Hard negative mining applied to lip-based biometric authentication
%\item Robustness to presentation attacks

\begin{keyword}
Lip-based biometric authentication \sep Siamese neural network \sep Hard-negative mining \sep Presentation attack detection \sep One-shot learning \sep GRID dataset



\end{keyword}

\end{frontmatter}

%\linenumbers

\section{Introduction}
\label{sec:introduction}

The ability to authenticate a person in order to provide access rights is a vital part of any security system with applications in public security, financial institutions, military, etc. Authentication, by definition, is a process of correctly determining a person’s identity. This can be done by using one or more of the following: something that a person has, something that a person knows, or something that a person is. The safest approach is using something that a person is, since what a person is can be difficult to steal or replicate. This is called biometric authentication \cite{jain_biometrics_2006} and is based upon a unique feature set of a person which differentiates it from others. Biometric features are grouped into two categories: physical and behavioral features. Physical feature is a characteristic inherently associated with a person. Face, fingerprint, iris, palm, etc. are examples of widely used physical features. A behavioral feature is an innate or practiced act or skill that a person expresses as a pattern of behavior. Voice, handwriting, body posture, etc. are typical behavioral features.

Face, as a physical biometric feature, gained popularity in the 1990s with the introduction of the eigenfaces approach \cite{turk_eigenfaces_1991} which projected face images onto a feature space that spans the significant variations among them for the purpose of face recognition. Problems in this space included sensibility under varying pose and illumination \cite{jain_biometrics_2006}. Early success was based on handcrafted local pattern methods \cite{chengjun_liu_gabor_2002,ahonen_face_2006} and machine learning \cite{cao_face_2010,phillips_support_1998}. Closing in on human-level performance was enabled by deep learning \cite{taigman_deepface_2014} and the availability of big and varied datasets \cite{deng_imagenet_2009,huang_labeled_2008}.

Even with exceptional recognition performance, face as a biometric feature lacks resistance to imposter attacks in the form of presentation attacks \cite{ramachandra_presentation_2018}. Presentation attacks are attempts to impersonate an enrolled user, in further text - client, in front of the authentication sensor in order to successfully authenticate. Most common presentation attacks are the print attack and video replay attack. A print attack involves presenting a printed picture of a registered person’s face to the authentication sensor. A video replay attack does the same for a video of a registered person being played on a screen in front of the sensor.

The existence of these attack methods requires secure systems, relying on facial features, to have the capability to detect them (presentation attack detection - PAD) or to confirm the genuine presence of a live person (liveness detection). Early methods against presentation attacks relied on handcrafted feature detectors which had a goal of detecting liveness in the form of blinking \cite{pan_eyeblink-based_2007,jee_liveness_2008}, head movement \cite{wang_face_2009}, gaze tracking \cite{bigun_assuring_2004}, etc. These methods proved successful in detecting print attacks but struggled with video replay attacks \cite{yu_deep_2021}. Progress against video replay attacks was made with the coming of specialized datasets that enabled the use of deep learning \cite{chingovska_effectiveness_2012,costa-pazo_replay-mobile_2016}, but the problem is still far from solved.

Lip-based biometric authentication (LBBA), a promising approach to robust biometric authentication, focuses on the mouth region of the face for two reasons. The first reason is the physical uniqueness with regard to the shape and imprint/texture of the lips \cite{wang_physiological_2012}. Different people have physically different lips - like a fingerprint, lips are unique to each person. In \cite{tsuchihashi_studies_1974} authors demonstrated this on a sample of 1364 people (757 men, 607 women) in Japan, aged from 3 to 60 years. It was shown that no pair of subjects had the same lip print. Additionally, the lip prints of the subjects were examined every month for three years in order to determine whether lip prints are permanent or not. No person showed any significant change during this period. The second reason is the behavioral uniqueness of the visual style of speech, i.e. the visual characteristics of the entire region around the mouth area, including the lips, position/visibility of teeth, oral cavity, and tongue over time \cite{wang_physiological_2012, singh_speaker_2012}. Different subjects in the pronunciation of the same phrase do not create identical visual sequences, from a video perspective \cite{benedikt_assessing_2010}. A phrase that is spoken for the purpose of LBBA, hereinafter, is referred to as an authentication phrase, while a spoken realization of a phrase by a particular speaker is referred to as an utterance.

Modern LBBA solutions use one-shot deep learning to analyze both physical and behavioral features for an easy enrollment procedure. Research in \cite{bebis_one-shot-learning_2019} demonstrated the first use of one-shot learning for LBBA using a siamese neural network and a single phrase from the XM2VTS dataset in a closed-set protocol setting. This approach was extended to multiple phrases in an open-set protocol setting with two additional datasets in \cite{wright_understanding_2020}. All of these approaches model behavioral characteristics only as style-of-speech, without relation to what is being said. This is reflected in the selection of training pairs, where positive pairs are of the same speaker and negative pairs are of different speakers. A flaw of this approach is that the system is not resistant to video replay attacks, in which a client could speak any phrase and still be authenticated. In this case, it is also impossible to change the authentication key, here corresponding to the physical and behavioral feature embedding. Additionally, the use of only different speakers in negative pairs makes it difficult to evaluate the efficiency of behavioral features and carries the risk of overfitting to physical ones.
 
In this paper, we propose a novel approach to one-shot LBBA that addresses these issues for the purpose of improving robustness against video replay attack. We do this by modeling behavioral features to discriminate against what is being said, in the form of corresponding lip movements, in addition to style-of-speech. To the best of our knwoledge, this is the first time phrase information has been used in one-shot deep learning LBBA. First, we customize the GRID dataset to obtain the right diversity and quantity of pairs needed to train a siamese neural network for our task. Negative pairs are defined as either pairs of different speakers, pairs of a different phrase being said, or both. Then we train a siamese neural network based on 3D convolutions and recurrent neural network layers for the task of creating an input embedding compatible with a cosine similarity function. For efficient learning of the underlying representation, we use a custom triplet loss which creates negative pairs on the fly and performs hard-negative mining across a batch.

The paper is structured in the following way. In Section~\ref{sec:related_work}, a brief overview of lip-based biometric identification approaches is given. The proposed approach for lip-based biometric authentication that integrates physical as well as behavioral person characteristics is presented in Section~\ref{sec:proposed_approach}. The obtained results along with the accompanying discussion are given in Section~\ref{sec:results}. In the end, a conclusion is given.

\section{Related work}
\label{sec:related_work}

Early work in \cite{luettin_speaker_1996}, based on spatio-temporal analysis of a talking face, performed speaker identification. Visual lip boundary and intensity parameters were used as features to train Hidden Markov models (HMMs) with mixtures of Gaussians as classifiers. Performance on the limited Tulips1 dataset \cite{movellan_visual_1994}, which includes 12 speakers saying the first 4 digits in English language, achieved around 90\% accuracy, depending on the setting. 

Evaluation on bigger and more diverse datasets was made possible by the introduction of the M2VTS dataset \cite{goos_m2vts_1997} and its extended version - XM2VTS \cite{messer_xm2vtsdb_2000}. The XM2VTS dataset is a multi-modal dataset consisting of frontal face recordings of 295 subjects created for the purpose of developing personal identity verification systems. Dataset is collected in 4 sessions over a period of 5 months to allow for a change in physical appearance. Each session includes a "speech-shot" for each speaker, where a subject is recorded speaking two digit-only phrases and a phonetically balanced phrase. Each phrase is recorded two times in a session.

In \cite{goos_evaluation_2003} authors used M2VTS to perform speech and text-dependent speaker recognition. The authors found that different features are effective for each task. Static features, such as raw-pixel information, proved sufficient for speaker recognition while features of a temporal nature showed better performance for the task of speech recognition. This was confirmed in \cite{cetingul_discriminative_2006} where, using HMMs, different lip motion features were explored on the same recognition problem using the MVGL-AVD database. A two-stage discrimination analysis technique involving spatial Bayesian feature selection and temporal Linear Discriminant Analysis (LDA) was proposed and found to improve performance in both speaker identification and speech-reading.

Even with improving accuracy, lip features still haven’t been shown to include unique speaker information for use in biometric systems. Research regarding LBBA continued by exploring the use of lip features jointly with audio \cite{faraj_motion_2006,faraj_audiovisual_2007,chetty_biometric_2009}. In \cite{faraj_motion_2006} authors used orientation estimation in 2D manifolds on the XM2VTS dataset to compare audio and visual features for speaker identification. Audio-visual features were shown to perform the best with a 98\% test set accuracy, while visual-only features achieved only 78\% in speaker identification. The same approach was explored in \cite{faraj_audiovisual_2007}, in the context of presentation attack and liveness detection. Authors found a way to exploit lip-motion features, without requiring segmentation or lip tracking, to achieve verification rates up to 94\%.

A first detailed analysis of the discriminative power of LBBA was done in \cite{benedikt_assessing_2010} where authors explored the uniqueness and permanence of lip movements, and facial actions as a whole. Analysis was made with two 3D cameras at 48 frames per second (FPS) on a proprietary dataset of short and isolated verbal actions, created over a period ranging from several weeks to several months, depending on the subject. Additional care was put into selecting and evaluating phrases depending on their phoneme/viseme structure. Phonemes are the smallest sound units that can be distinguished and that together make up a spoken word. Viseme is the visual equivalent of a phoneme, i.e. the smallest distinguishable visual unit created during pronunciation. Several phonemes can belong to a single viseme \cite{cox_confusability_2004} which can cause different spoken phrases to be presented with the same viseme sequence. The authors emphasize that this needs to be taken into account when selecting phrases. To extract features for analysis, the authors propose a recognition system using facial dynamics, including data preprocessing techniques, a feature extraction method, and a pattern-matching algorithm derived from dynamic time warping (DTW). In their evaluations, they find that even short verbal facial actions provide sufficient information for person identification. A hierarchy in the reproducibility of different types of phrases is also observed, where short vowels were shown to be more repeatable over long periods of time than long vowels. Repeatability in this context is defined as the ability to reproduce the same sequence of visual features for facial action over time. A drawback of all the aforementioned approaches was the need for repeated calibration and training of models for enrolling each additional phrase or speaker.

With the advancement of deep learning technology, analysis of lip dynamics was further applied to lip reading \cite{chung_lip_2017, shillingford_large-scale_2018,wand_lipreading_2016,assael_lipnet_2016} on the GRID \cite{cooke_audio-visual_2006} dataset. A comparison of a deep learning approach, using LSTMs, to a conventional SVM approach was done in \cite{wand_lipreading_2016}. Authors performed visual lip reading by performing word classification and showed 11.6\% improvement over feature-based methods. A study of the GRID dataset, recorded at 25 frames per second, showed that letters occupied 3-4 frames on average, while longer words occupied more than 10 frames.

In \cite{assael_lipnet_2016} authors presented LipNet an end-to-end sentence level lip-reading architecture based on 3D convolutions and RNN’s which was able to successfully recognize viseme sequences along with context, from the mouth region video during speech, and output a distribution over a sequence of character tokens. Unseen speaker generalization was reported at 88.6\% sentence-level word accuracy. LipNet architecture, even though specialized for lip reading, inspired one-shot solutions to LBBA \cite{bebis_one-shot-learning_2019,wright_understanding_2020}. This meant that only a single prior example is needed for enrolling new users.

In \cite{bebis_one-shot-learning_2019}, the first one-shot approach to modeling both the physical and behavioral features for the purpose of LBBA was proposed. Authors trained a siamese neural network named LipAuth, inspired by LipNet \cite{assael_lipnet_2016}, with a triplet loss function using RGB video recordings of a person’s mouth region during speech. Dataset consisted of a single 20 digit phrase originating from the XM2VTS dataset, repeated 8 times by each speaker. Closed-protocol evaluation is performed with anchor subjects originating from the training set. The reported Equal Error Rate (\textit{EER}) on the evaluation set is 0.93\%. Since the work only uses a unique phrase for training the model, a negative pair is modeled only as a pair of utterances of the same phrase by different speakers. This raises concerns about does the model need to understand any behavioral features for authenticating the subjects as it can overfit on physical appearance features for discriminating subjects. If this is the case it may pose a risk of image print and video replay attacks, which hasn’t been explored in this paper.

In contrast, we use a set of 64 authentication phrases derived from a customized GRID dataset and model negative pairs as either pairs of utterances belonging to different speakers or/and different authentication phrases. Our dataset split doesn't share speakers across sets and our test set evaluation was done using a threshold that matches the \textit{EER} threshold on the training set rather than measuring \textit{EER} directly on the test set.

To gain insight into the generalization capabilities of the proposed approach in the scenario of multiple phrases, authors of \cite{bebis_one-shot-learning_2019} extended their approach in \cite{wright_understanding_2020} with an open-set protocol on the XM2VTS dataset and two additional private datasets. The first additional dataset, qface, contains only digit phrases with up to 20 digits per phrase. It was used to test the authentication accuracy influence of phrase lengths, phrase differences between enrollment and authentication, and cross-dataset enrollment and authentication. The second additional dataset, FAVLIPs was collected over 4 months and its aim was to test the LBBAs performance on mobile devices under varied conditions. On XM2VTS dataset authors achieved a test set \textit{EER} of 1.65\%. Their approach models the behavioral features on all datasets as style-of-speech without relation to what is being said, i.e. authors expected the enrolled user to authenticate even if he said a phrase that differs from his enrollment phrase. This is reflected in the selection of triplet pairs where achor-positive pairs are of the same speaker and anchor-negative pairs are of different ones. This authentication method has a flaw of being vulnerable to video replay attacks. A video of a client speaking any phrase has the potential to successfully authenticate and therefore is a security threat.

In our approach, we model the behavioral features as both lip movements corresponding to the authentication phrase and the speaker’s style of speech. This way, in order to authenticate, the right subject needs to say the right authentication phrase. This approach is more robust to video replay attacks as the attacker would need the video of the enrolled person saying its exact authentication phrase to spoof the system.

\section{Proposed approach to one-shot learning for LBBA}
\label{sec:proposed_approach}

In this section, a novel approach for one-shot LBBA is presented. Our method’s aim is to advance one-shot LBBA by extending behavioral features to enable discriminating against what is being said in addition to style-of-speech. The intent is to significantly improve the resistance to video replay attacks of the client speaking. We model this problem as training a siamese neural network to predict the similarity of inputs represented as videos of speech of the subject's mouth region. The possible network inputs and their labels, as per our aim, are given in \autoref{tab:network_inputs}.

\begin{table}[b]
	\caption{Possible types of siamese neural network input pairs and their labels for the goal of authenticating only when the same phrase is spoken by the same speaker.}
	\label{tab:network_inputs}
	\footnotesize
	\begin{tabularx}{\textwidth}{>{\hsize=0.5\hsize}X
                              >{\hsize=2.\hsize}X
                              >{\hsize=0.5\hsize}X}
	\toprule
	%\rowcolor{atomictangerine}[\dimexpr\tabcolsep + 1pt\relax]
	\textbf{Type ID}	& \textbf{Description} &  \textbf{Label} \\
    \midrule
        {Type 1} 	&	Pair of same speakers saying the same phrase		& positive \\
        \midrule
		{Type 2} 	&	Pair of same speakers saying different phrases		& negative \\
		\midrule
		{Type 3} 	&	Pair of different speakers saying the same phrase 	& negative \\
		\midrule
		{Type 4} 	&	Pair of different speakers saying different phrases	& negative \\
	\bottomrule
	\end{tabularx}
\end{table}

To create a dataset with adequate diversity and quantity of such pairs we resorted to customizing the GRID dataset. Our siamese network is constructed from a LipNet inspired backbone and a custom head. Loss was calculated using a custom triplet loss function which inherently performs customizable hard-negative mining across a batch.

\subsection{Customized GRID dataset}
\label{subsec:modified_grid}

Open-source and even private datasets used for LBBA lack the quantity and diversity of pairs shown in \autoref{tab:network_inputs}. The widely used XM2VTS dataset contains only three unique phrases being spoken 8 times by each speaker. Included phrases are also relatively long which does not suit the problem at hand as shorter ones are preferred \cite{benedikt_assessing_2010}. This makes evaluating the model’s performance on diverse authentication phrases difficult. We solve the aforementioned problems by adapting the GRID dataset to provide adequate speaker and authentication phrase diversity. 

GRID dataset contains audio and video recordings of subjects uttering phrases that follow a strict pattern. Videos were made using a Canon XM2 camcorder at 25 FPS and a resolution of 720x576 px. Phrases follow a pattern: $"\textless command:4\textgreater\textless color:4\textgreater\textless preposition:4\textgreater\textless letter:25\textgreater\textless digit:10\textgreater\textless adverb:4\textgreater"$ with the word category written before the colon and number of choices for the word category written after the colon. Phrase pattern along with all possible word category choices is shown in \autoref{tab:grid_prompts}. 

\begin{figure}[htb]
	\centering
	\begin{subfigure}[t]{.45\linewidth}
  		\includegraphics[width=\linewidth]{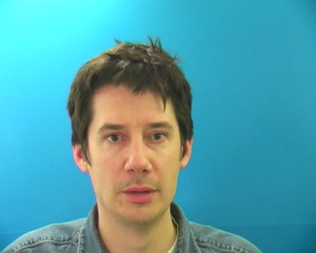}
  		\caption{Speaker 1}
  	\end{subfigure}
   	\begin{subfigure}[t]{.45\linewidth}
  	  	\includegraphics[width=\linewidth]{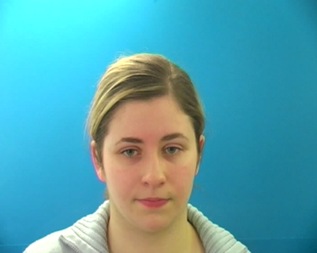}
  		\caption{Speaker 11}
  	\end{subfigure}
	\caption{Video frame examples from GRID dataset.}
	\label{fig:grid_example} 
\end{figure}

An example of a GRID phrase is "place blue at F 9 now". It can be seen that the GRID dataset contains phrases that often only differ by one word, making it more challenging to model behavioral features that can discriminate between the phrases compared to the XM2VTS dataset. The dataset contains 33 subjects (15 female and 18 male), identified by their ID number, each repeating a unique phrase resulting in 33000 video recordings of utterances. Recordings are covering the subject’s frontal face and shoulder area in front of a blue background as shown in \autoref{fig:grid_example} where two frames from the GRID dataset are shown. 

\begin{table}[htb]
\centering
	\caption{GRID dataset phrase pattern and available word choices for each word category.}
	\label{tab:grid_prompts}
	\footnotesize
	\begin{tabularx}{\textwidth}{XXXXXX}
	%\rowcolor{atomictangerine}[\dimexpr\tabcolsep + 1pt\relax]
	\toprule
	\textbf{command}			& \textbf{color}		& \textbf{preposition}		& \textbf{letter}	& \textbf{digit}	& \textbf{adverb} \\
    \midrule
        bin 				& blue					& at						& A-Z				& 1-9, zero 		& again \\
		lay 				& green					& by						& excl. W			&					& now \\
		place 				& red					& in						& 					&					& please \\
		set 				& white					& with						& 					&					& soon \\
	\bottomrule
	\end{tabularx}
\end{table}

On its own, the GRID dataset has unique phrases per speaker. But, even though phrases are unique, they follow a strict pattern in which certain sub-phrases, belonging to a sub-pattern, are spoken multiple times both by a single speaker and across speakers. Accompanying that with the dataset’s provided word alignments we are able to cut video recordings to reflect the sub-pattern. There are multiple choices of sub-patterns to choose from. In \autoref{tab:possible_phrases} we compare some of the considered choices. We chose the "command, color, preposition" sub-pattern as it provides 64 possible authentication phrases with a median of 16 utterances of the same authentication phrase per speaker. The resulting cut video recordings are further processed for model training.

\begin{table}[tb]
    \caption{Properties of different sub-patterns from the GRID dataset.}
    \label{tab:possible_phrases}
    \footnotesize
    \begin{tabularx}{\linewidth}{>{\hsize=1.1\hsize}X
                              >{\hsize=0.6\hsize}Y
                              >{\hsize=0.6\hsize}Y
                              >{\hsize=0.6\hsize}Y
                              >{\hsize=0.6\hsize}Y}
	\toprule
	& \multicolumn{4}{c}{\textbf{Sub-pattern}} \\
	\cline{2-5}
    & command, color, preposition & letter, digit, adverb & digit adverb & command, color, preposition, letter\\
    \midrule
    Number of speakers																& 33		& 33 		& 33		& 33 \\
    \midrule
    Number of phrases																& 64		& 1,000 	& 40		& 1,150 \\
    \midrule
    Median of counts of utterances per phrase										& 510		& 33 		& 825		& 36 \\
    \midrule
    Median of counts of utterances per phrase by a single speaker					& 16		& 2			& 50		& 4 \\
    \midrule
    Number of unique speaker and phrase pairs										& 2,112		& 16,500	& 660		& 10,700 \\
    \midrule
    Maximum number of possible positive pairs (Type 1 pairs)						& 258,920	& 33,000	& 825,000	& 57,800 \\
    \bottomrule
    \end{tabularx}
\end{table}

\subsection{Dataset preprocessing}
\label{subsec:dataset_preprocessing}

\begin{figure}[tbh!]
	\centering
  	\includegraphics[width=0.65\linewidth]{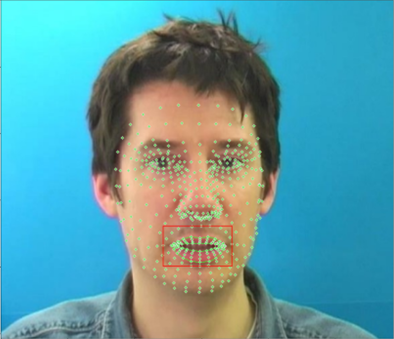}  
	\caption{An example of the Face Mesh landmark mask superimposed on a frame of a speaking face from the GRID dataset.}
	\label{fig:face_mesh} 
\end{figure}

\begin{figure}[tbh!]
	\centering
  	\includegraphics[width=0.65\linewidth]{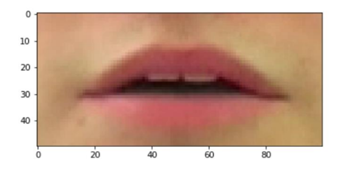}  
	\caption{An example of an image of the mouth region, which has been obtained by cropping a speaker's face image.}
	\label{fig:mouth_region} 
\end{figure}

Video recordings, cut to represent the chosen subpattern, are further cropped to only contain the mouth region. Cropping is done in two stages. First,  MediaPipe Face Mesh, an open-source pretrained model \cite{lugaresi_mediapipe_2019} based on MobileNetV2, is used to infer 468 face landmarks, most notably lip region landmarks, for each video frame. An example of the face landmark mask on a face from the GRID dataset is shown in \autoref{fig:face_mesh}. After landmark detection, a rectangle shaped mouth region defined by its side midpoints, corresponding to landmark indices 57, 287, 164, and 18, is cropped. An example of a cropped image is shown in \autoref{fig:mouth_region}. The resulting image is resized to 100x50px and converted to a grayscale image. Video recordings that included frames where face landmark detection confidence was below 50\% were discarded, which amounted to approximately 4\% of all video recordings.

\subsection{Model architecture}
\label{subsec:model_architecture}

Siamese neural networks are used to learn a similarity function between distinct network inputs. Architecture is characterized by identical network branches with shared weights where each outputs a vector embedding of a particular input. The objective is to reduce the embedding distance between pairs of inputs of the same label (positive pairs) and increase it between pairs of inputs of different labels (negative pairs). This is done by applying a distance metric to the output embeddings and using a ranking loss to train the network with backpropagation. Since the network learns an embedding function for the input it is able to perform one-shot learning which is particularly useful in biometric applications as enrolment only requires one example.

Model architecture used in this paper consists of a siamese neural network with two branches where each has a network backbone inspired by LipNet, similar to \cite{bebis_one-shot-learning_2019,wright_understanding_2020}, and a custom head. Original LipNet architecture started with three layers of 3D convolutions along with 3D max-pooling and 3D dropout. Following it were two bidirectional GRU layers feeding to a final linear layer and softmax. Input to the network consisted of sequences of RGB frames of a person’s mouth region while speaking. Output was a distribution over a set of phonemes for the task of lip reading. 

In this work, the architecture is repurposed to produce an input embedding. The 3D convolutional layers are adapted to process grayscale image sequences and the bi-directional GRU layers are modified accordingly. Hidden states of the final bi-GRU layer are fed to a custom head which calculates mean and max of the hidden states across the time-axis and concatenates the result to be fed to a final linear layer which outputs 256-dimensional input embeddings. Output is L2 normalized. The proposed neural network architecture is given in \autoref{tab:nn_structure}, where B denotes batch size, C denotes number of channels, T denotes time, H denotes frame height, W denotes frame width, F and G denote feature dimensions and E denotes the embedding dimension.

\begin{table}[htb!]
    \caption{Proposed model architecture.}
    \label{tab:nn_structure}
    \footnotesize
    \begin{tabularx}{\linewidth}{>{\hsize=0.2\hsize}X
    							>{\hsize=0.27\hsize}X
                              	>{\hsize=0.28\hsize}X
                              	>{\hsize=0.25\hsize}X}
    \toprule
	\textbf{Layer}		& \textbf{Size/Stride/Pad}		& \textbf{Input Size}		& \textbf{Dimension Order}	\\	\midrule
	Conv3D				& (3,5,5)/(1,2,2)/(1,2,2)		& -1 x 1 x 50 x 100 x 50	& B, C, T, H, W				\\	\midrule
	Relu				& -								& -1 x 32 x 50 x 50 x 25	& B, C, T, H, W				\\	\midrule
	MaxPool3D			& (1,2,2)/(1,2,2)				& -1 x 32 x 50 x 50 x 25	& B, C, T, H, W				\\	\midrule
	Conv3D				& (3,5,5)/(1,1,1)/(1,2,2)		& -1 x 32 x 50 x 25 x 12	& B, C, T, H, W				\\	\midrule
	Relu				& -								& -1 x 64 x 50 x 25 x 12	& B, C, T, H, W				\\	\midrule
	MaxPool3D			& (1,2,2)/(1,2,2)				& -1 x 64 x 50 x 25 x 12	& B, C, T, H, W				\\	\midrule
	Conv3D				& (3,3,3)/(1,1,1)/(1,1,1)		& -1 x 64 x 50 x 12 x 6		& B, C, T, H, W				\\	\midrule
	Relu				& -								& -1 x 96 x 50 x 12 x 6		& B, C, T, H, W				\\	\midrule
	MaxPool3D			& (1,2,2)/(1,2,2)				& -1 x 96 x 50 x 12 x 6		& B, C, T, H, W				\\	\midrule
	Bi-GRU				& 1728							& 50 x -1 x 1728			& (T , B , C * H * W)		\\	\midrule
	Bi-GRU				& 256							& 50 x -1 x 256				& (T , B , F)				\\	\midrule
	CustomAvgMax		& -								& 50 x -1 x 128				& (T , B , G)				\\	\midrule
	Linear				& 256							& 50 x -1 x 128				& (T , B , G)				\\	\midrule
	L2-Normalize		& -								& -1 x 256					& (B, E)					\\
	\bottomrule
	\end{tabularx}
\end{table}

\subsection{Hard-negative mining and the modified triplet loss function}
\label{subsec:hard_heative}

One challenge of training siamese neural networks is the issue of imbalanced training data.  When the number of negative pairs far outweighs the number of positive pairs, the network may develop a bias towards correctly classifying negative pairs while struggling with positive pairs. 
Additionally, negative pairs need to be sufficiently difficult in order for learning to occur. In \cite{schroff_facenet_2015} it was demonstrated that increasing the difficulty of the negative pairs during training can improve the network's overall performance. This technique is called hard-negative mining. In this work, hard-negative mining is performed across a batch inside a custom triplet loss function. Batch-wise mining is chosen because of its speed, in comparison to mining over the whole dataset. Our triplet loss accepts only positive pairs as input and generates negative pairs on the fly.

Let $A=\{a_i:i=1,...,64\}$ be the set of authentication phrases which belong to a chosen sub-pattern from GRID dataset (e.g. $a_1=\textnormal{"bin blue at"}$). Let $S=\{s_j:j=1,...,33\}$ be the set of speakers from GRID dataset. Let $U_{i,j}=\{u_{i,j,k}:k=1,...,K_{i,j}\}$ be the set of utterances of authentication phrase $a_i$ spoken by speaker $s_j$ where $K_{i,j}$ is the set size. The unordered tuple $(u_{m,n,l},u_{o,p,q})$ is said to be a positive pair if $m=o$, $n=p$ and $l \neq q$. If $m \neq o$ and/or $n \neq p$, then the tuple is a negative pair.

The prerequisite for proper operation of our loss function are batches where each positive pair inside a batch contains utterances sampled from a different utterance set $U_{i,j}$. This in turn makes utterances belonging to two different positive pairs form a negative pair. This symmetry is used in the loss function to dynamically create negatives and perform hard-negative mining.

A batch of size $N$ is passed to the network as two input tensors $\mathbf{X_1}$ and $\mathbf{X_2}$ where $\mathbf{X_1}$ is the input tensor containing pair elements that enter the first branch and $\mathbf{X_2}$ containing pair elements that enter the second branch. The outputs of the neural network are two embedding matrices $\mathbf{Z_1}$ and $\mathbf{Z_2}$ of size $(N \times 256)$ where each row represents an L2 normalized input embedding for the corresponding pair element. 

Inside the loss function, a matrix multiplication of $\mathbf{Z_1}$ and $\mathbf{Z_2}$ is performed according to \autoref{eq1} in order to obtain a similarity matrix. The entries in this matrix represent the cosine similarity between pairs of elements. Because of our batch constraint, positive pair scores are located on the main diagonal while dynamically created negative pair scores are off-diagonal.

\begin{equation}	\label{eq1}
	\mathbf{S} = \mathbf{Z}_{1} \times \mathbf{Z}_{2}^{T}
\end{equation}

Next, the loss is calculated in two parts. $\mathbf{loss_1}$ puts emphasis on the mean and $\mathbf{loss_2}$ on the maximum of negative scores for each row according to \autoref{eq2}. This design was chosen to allow for adjusting the influence of the hardest negatives and negatives as a whole to the loss function. The margin $m$ for both losses is set to 0.5.  

\begin{equation}	\label{eq2}
	\begin{split}
	\mathbf{Loss}(\mathbf{loss_1},\mathbf{loss_2}) &= \textnormal{mean}(0.5*\mathbf{loss_1}+0.5*\mathbf{loss_2}) \\
	[\mathbf{loss_1}]_{i} &= \max(0,m-\mathbf{p}_{i} + \mathbf{maxN}_{i}) 	\\
	[\mathbf{loss_2}]_{i} &= \max(0,m-\mathbf{p}_{i} + \mathbf{meanN}_{i})	\\
	[\mathbf{maxN}]_i &= \max((\mathbf{S}-\mathbf{I}_{N})_{i}) \\
	[\mathbf{meanN}]_i &= \frac{\sum_{j=1}^{N} (\mathbf{S} - \mathbf{I}_{N})_{ij}} {N - 1} \\
	\mathbf{p} &= \textnormal{diag}(\mathbf{S})  \\
	\end{split}	
\end{equation}

\section{Results and discussion}
\label{sec:results}

\subsection{Experimental setup}
\label{subsec:experimental_setup}

Experiments are performed using an open-set protocol on the customized GRID dataset. Out of 33 speakers, 25 were chosen for the training, 4 for validation, and 4 for the test set. Gender was taken into account when selecting speakers for equal representation in the validation and test set. \autoref{tab:data_split_speakers} shows speakers in each split with the color showing gender (blue for male, red for female).

\begin{table}[tb]
    \caption{Distribution of speakers in each data split of the customized GRID dataset. Blue-colored IDs identify male speakers, while red-colored IDs identify female speakers.}
    \label{tab:data_split_speakers}
    \footnotesize
    \begin{tabularx}{\textwidth}{>{\hsize=1.1\hsize}Y
                              >{\hsize=0.45\hsize}Y
                              >{\hsize=0.45\hsize}Y}
    \hline
    \textbf{Training dataset}	&	\textbf{Validation dataset}	&	\textbf{Test dataset} \\
    \hline
    \textcolor{blue}{3}, \textcolor{red}{4}, \textcolor{blue}{6}, \textcolor{blue}{6}, \textcolor{red}{7}, \textcolor{blue}{8}, \textcolor{blue}{9}, \textcolor{blue}{10}, \textcolor{red}{11}, \textcolor{blue}{12}, \textcolor{blue}{13}, \textcolor{blue}{14}, \textcolor{red}{15}, \textcolor{red}{23}, \textcolor{red}{24}, \textcolor{red}{25}, \textcolor{blue}{26}, \textcolor{blue}{27}, \textcolor{blue}{28}, \textcolor{red}{29}, \textcolor{blue}{30}, \textcolor{red}{31}, \textcolor{blue}{32}, \textcolor{red}{33}, \textcolor{red}{34} 	& \textcolor{red}{16}, \textcolor{blue}{17}, \textcolor{red}{18}, \textcolor{blue}{19}	& \textcolor{blue}{1}, \textcolor{blue}{2}, \textcolor{red}{20}, \textcolor{red}{22}\\
    \hline    
	\end{tabularx}
\end{table}

Each split was then constructed by sampling only positive pairs (same speaker, same phrase) without repetition. 100,000 pairs were sampled for training, 10,000 for validation, and 10,000 for the test set. Training, evaluation and test set batch sizes are set to 80, 40, and 40, respectively. A validation set was used for evaluating hyperparameters before testing on the test set. If we consider the negative pairs that are implicitly created and evaluated by the loss function, the resulting number of pairs for the validation and test sets is shown in \autoref{tab:data_split_samples}, sorted by their type.

Training is done for 15 epochs using the Adam optimizer with a learning rate set to 10e-4. Dropout is set to zero. Several augmentation methods are applied similar to \cite{assael_lipnet_2016}. Deletion and duplication of video frames is set to a per-frame probability of 0.1. Random rotation of maximum 20 degrees to a clockwise or counterclockwise direction is performed with a per-video probability of 0.15. Random horizontal flip of all frames for a video is done with a probability 0.2.

\begin{table}[tb]
    \caption{Distribution of pair types inside the validation and test set of the customized GRID dataset.}
    \label{tab:data_split_samples}
    \footnotesize
    \begin{tabularx}{\textwidth}{>{\hsize=0.9\hsize}X
                              >{\hsize=0.6\hsize}X
                              >{\hsize=0.5\hsize}X}
    \toprule
    \textbf{Pair type}										&	\textbf{Validation dataset}	&	\textbf{Test dataset} \\
    \midrule
	Type 1: Positive, same speaker, same phrase    			& 10,000 ($\sim$2.5\%)		& 10,000 ($\sim$2.5\%)		\\	\midrule
	Type 2: Negative, Same speaker, different phrase		& 96,580 ($\sim$24.15\%)	& 96,586 ($\sim$24.15\%)	\\	\midrule
	Type 3: Negative, different speaker, same phrase		& 4,554 ($\sim$1.14\%)		& 4,634 ($\sim$1.16\%)		\\	\midrule
	Type 4: Negative, different speaker, different phrase	& 288,866 ($\sim$72.21\%)	& 288,780 ($\sim$72.20\%)	\\
	\midrule
	\textbf{Total}									& \textbf{400,000}		& \textbf{400,000}	\\
    \bottomrule    
	\end{tabularx}
\end{table}

\subsection{Results}
\label{subsec:results_discussion}

The proposed model performance is measured using biometric evaluation metrics false acceptance rate (\textit{FAR}) and false rejection rate (\textit{FRR}) as given in \autoref{eq4}.

\begin{equation}	\label{eq4}
	\begin{split}
	FRR = \frac{FN}{TP + FN} \\
	FAR = \frac{FP}{TN + FP}
	\end{split}	
\end{equation}

\textit{TP} and \textit{TN} refer to the number of correctly predicted positive and negative pairs, respectively. \textit{FP} and \textit{FN} refer to the number of incorrectly predicted positive and negative pairs, respectively. Threshold applied to model output when calculating these metrics on the validation and test set equals the threshold of Equal Error Rate (\textit{EER}) on the training set. \textit{EER} achieved on the training set is 0.00346 at a threshold of 0.45. Performance on the validation and test sets is given in \autoref{tab:far_frr_results}.

\begin{table}[tb]
\centering
	\caption{\textit{FAR} and \textit{FRR} metrics on validation and test set using a threshold of 0.45.}
	\label{tab:far_frr_results}
	\footnotesize
	\begin{tabularx}{\textwidth}{XXX}
	\toprule
	\textbf{Dataset}			& \textbf{$FAR$}		& \textbf{$FRR$}	 	\\
    \midrule
	Validation set				& 2.93\%					& 2.32\%			\\
	Test set					& 3.23\%					& 3.86\%			\\
 	\bottomrule
	\end{tabularx}
\end{table}

\begin{figure}[tb]
	\centering
	\begin{subfigure}[t]{.467\linewidth}
  		\includegraphics[width=\linewidth]{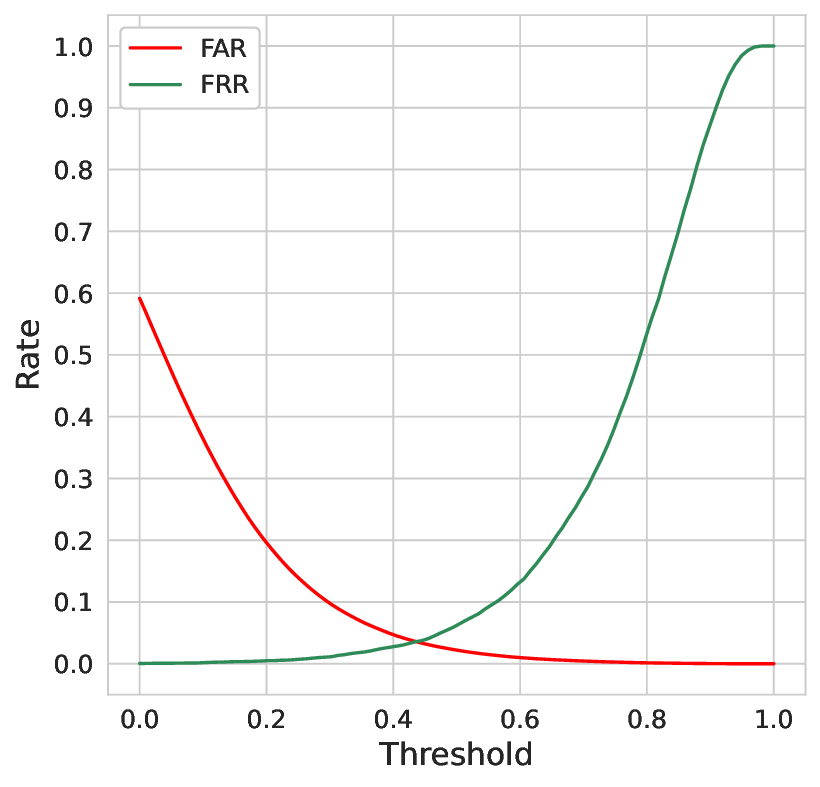}
  		\caption{\textit{FAR} and \textit{FRR} curves.}
  		\label{fig:main_results_a}
  	\end{subfigure}
   	\begin{subfigure}[t]{.493\linewidth}
  	  	\includegraphics[width=\linewidth]{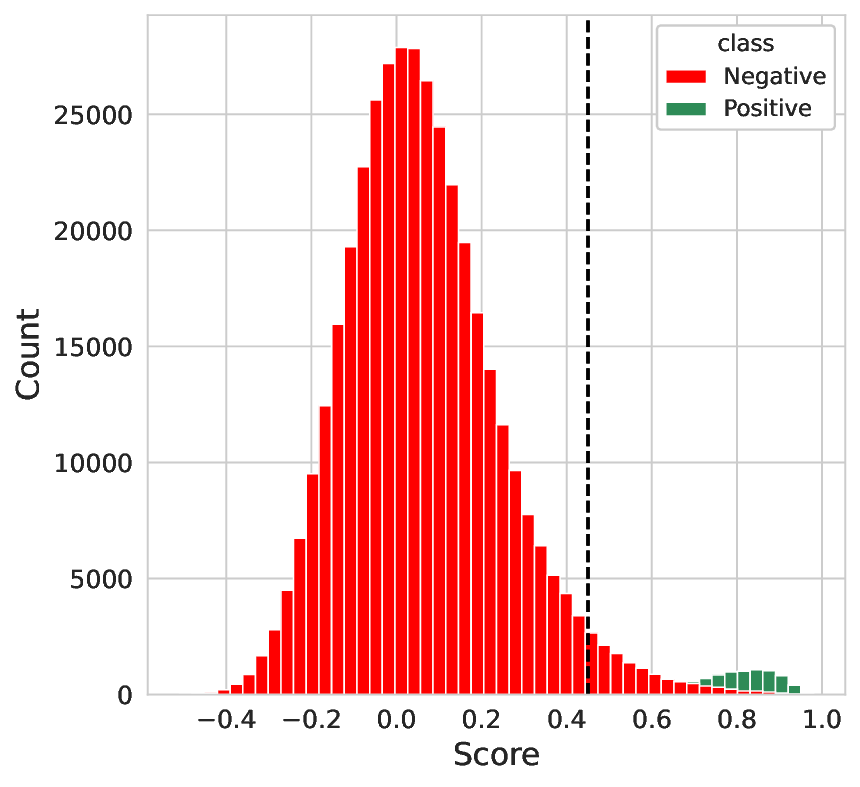}
  		\caption{Histograms of negative and positive pair prediction scores.}
  		\label{fig:main_results_b} 
  	\end{subfigure}
	\caption{Main results on the test set.}
	\label{fig:main_results} 
\end{figure}

Despite the added complexity of having to understand the spoken phrase, the model is able to achieve \textit{FAR} and \textit{FRR} comparable to SOTA one-shot deep learning approaches that rely only on style-of-speech \cite{wright_understanding_2020, bebis_one-shot-learning_2019}. This indicates that the model is able to offer much better protection against presentation attacks without sacrificing much on performance. \textit{FAR} and \textit{FRR} curves on the test set are given in \autoref{fig:main_results_a} and histograms of negative and positive pair scores are shown in \autoref{fig:main_results_b}. \textit{FAR} can be decreased by increasing the threshold, but at the expense of increasing \textit{FRR}.

\begin{figure}[tb]
	\centering
	\includegraphics[width=0.6\linewidth]{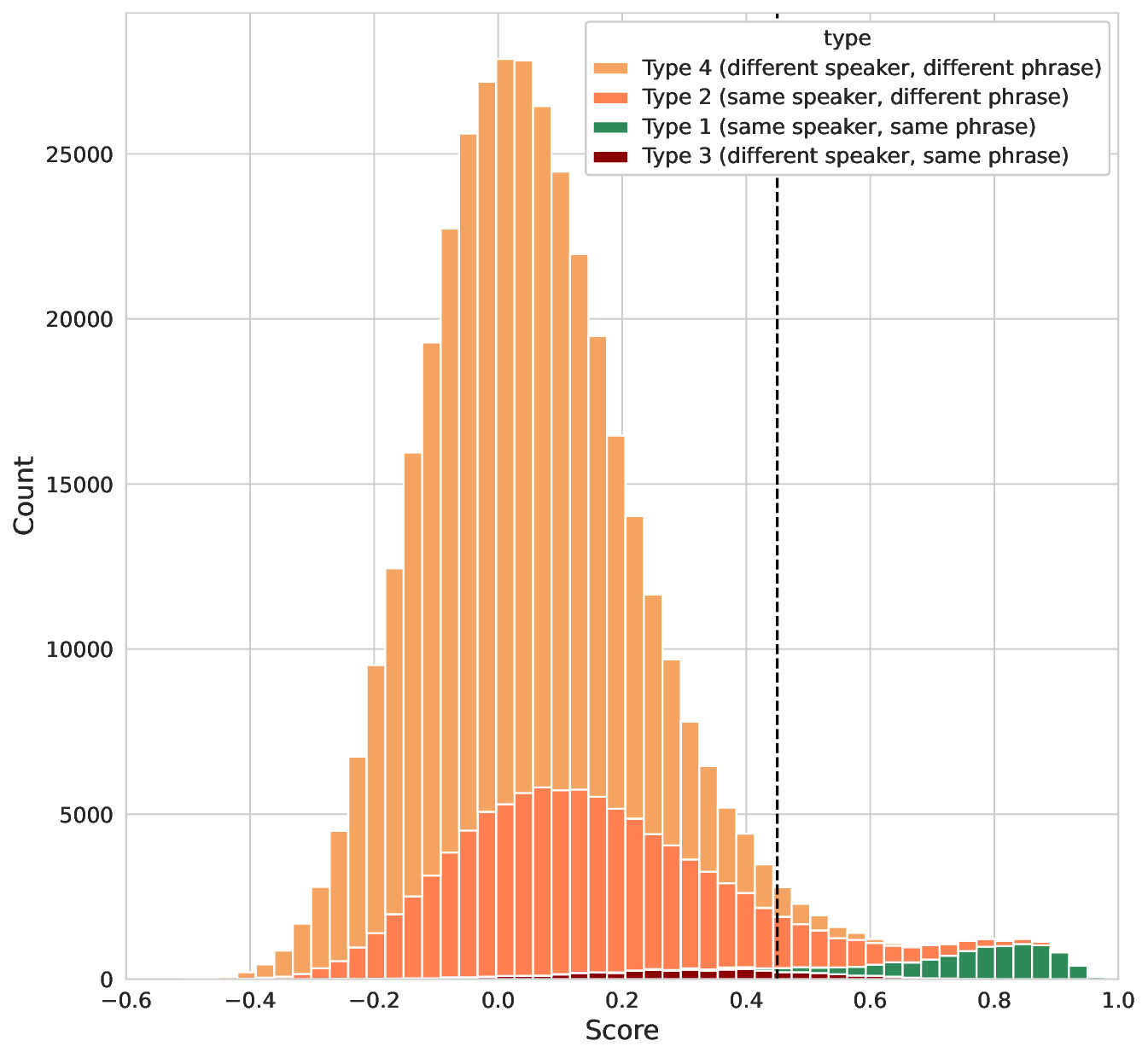}
	\caption{Stacked histograms of test set pair prediction scores, with respect to pair type.}
	\label{fig:histogram_scores} 
\end{figure}

\textit{FAR} metric in deep learning LBBA approaches is often reported without additional analysis, which does not provide insight into the impact of behavioral and physical features. To understand the impact of physical features on the model, we can analyze the performance on pairs where the physical features differ, which are pairs of type 3 (different speaker, same phrase). Similarly, we can examine the impact of behavioral features by looking at the performance on pairs where behavioral features differ, which are pairs of type 2 (different phrase, same speaker). To perform this analysis, score distributions for test set predictions are visualized using stacked histograms of each pair type. Histograms are shown in \autoref{fig:histogram_scores}, with the threshold of 0.45 indicated by a dashed vertical line.

\begin{figure}[h!]
	\centering
	\includegraphics[width=0.6\linewidth]{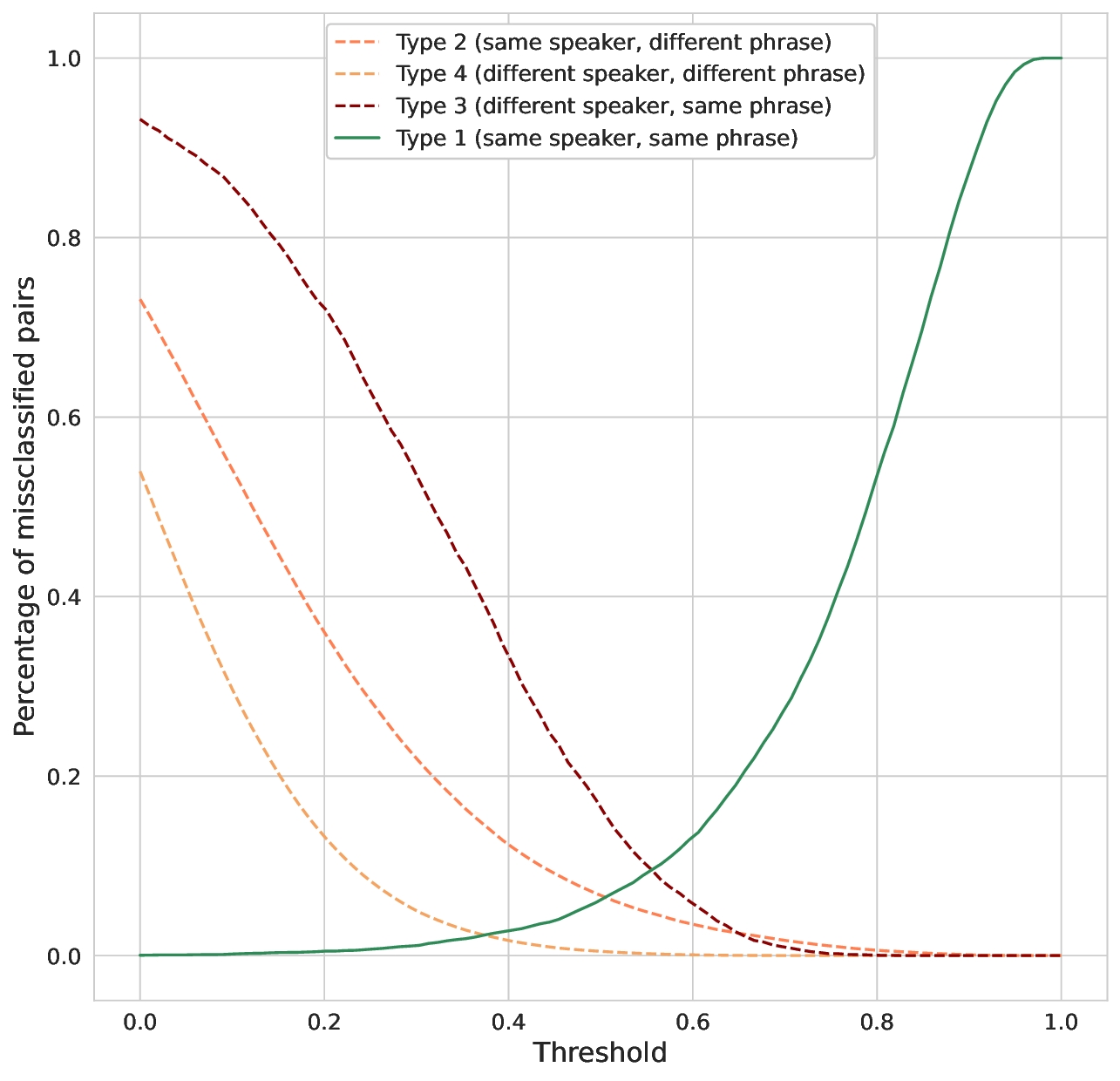}
	\caption{Percentage of missclassified pairs belonging to each pair type as the threshold is varied.}
	\label{fig:percentage_missclassified} 
\end{figure}

As expected, the model has the greatest confidence in its prediction when both physical and behavioral features are strictly the same or strictly different as shown for pairs of type 1 and type 2, respectively. On the other hand, the distributions for type 2 and type 3 pairs show that our model is better at discriminating based on behavioral features than physical ones. This effect can also be in part due to the imbalanced dataset in which only about 1\% of pairs are of type 3 compared to about 25\% for type 2. In \autoref{fig:histogram_scores_separated} and \autoref{fig:histogram_scores_separated_normalized}, for better visibility, the distributions for each type are separated and shown in their own plots along with normalized versions.

To get a more clear view of model performance on each type, \autoref{fig:percentage_missclassified} shows the percentage of misclassified pairs among all pairs of a specific type as the threshold is varied. This more clearly demonstrates the previously mentioned conclusions.

\begin{figure}[tbh]
	\centering
	\includegraphics[width=\linewidth]{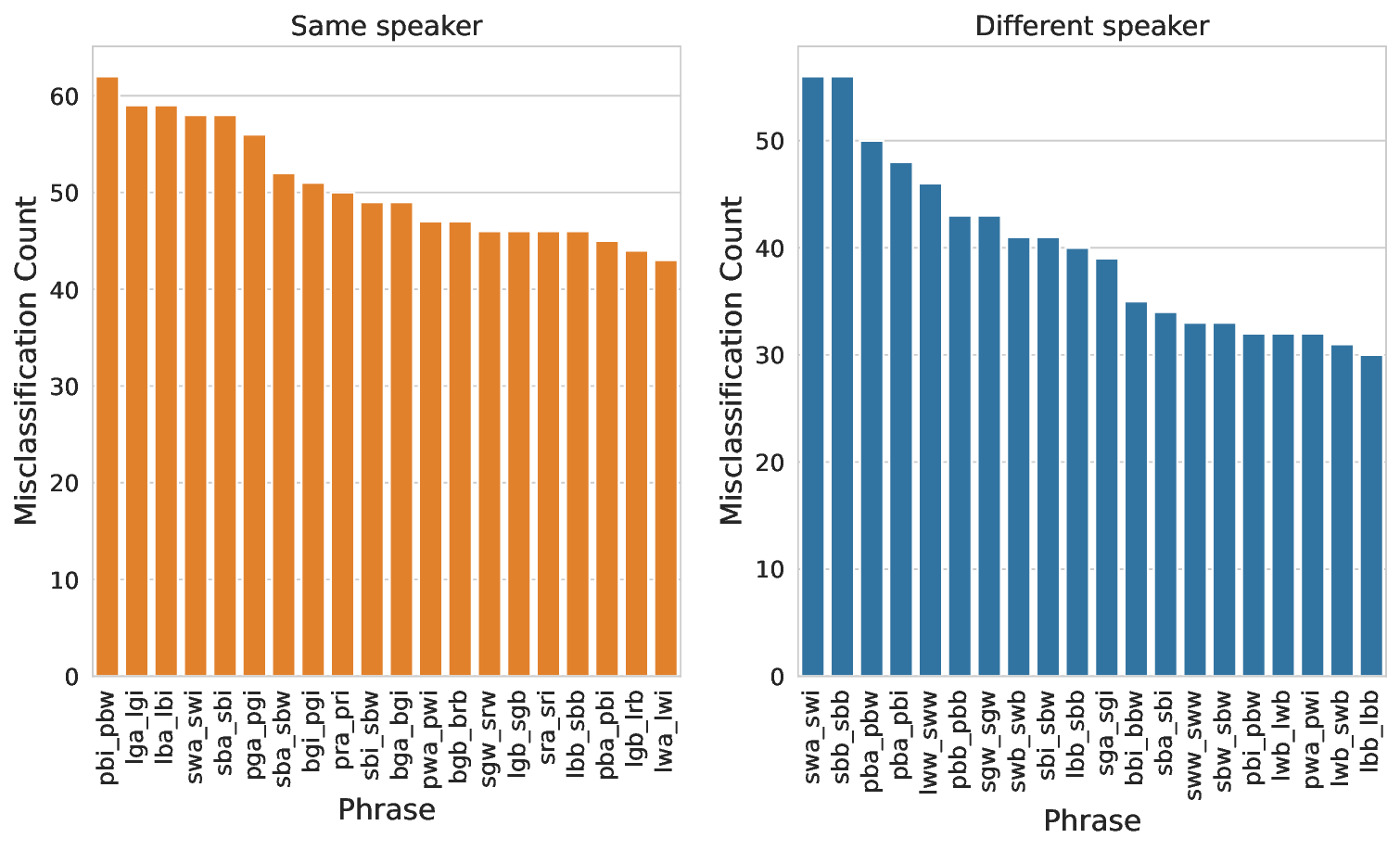}
	\caption{Top 20 most confused test set phrase pairs.}
	\label{fig:most_confused_phrases} 
\end{figure}

\begin{figure}[tbh]
	\centering
	\includegraphics[width=\linewidth]{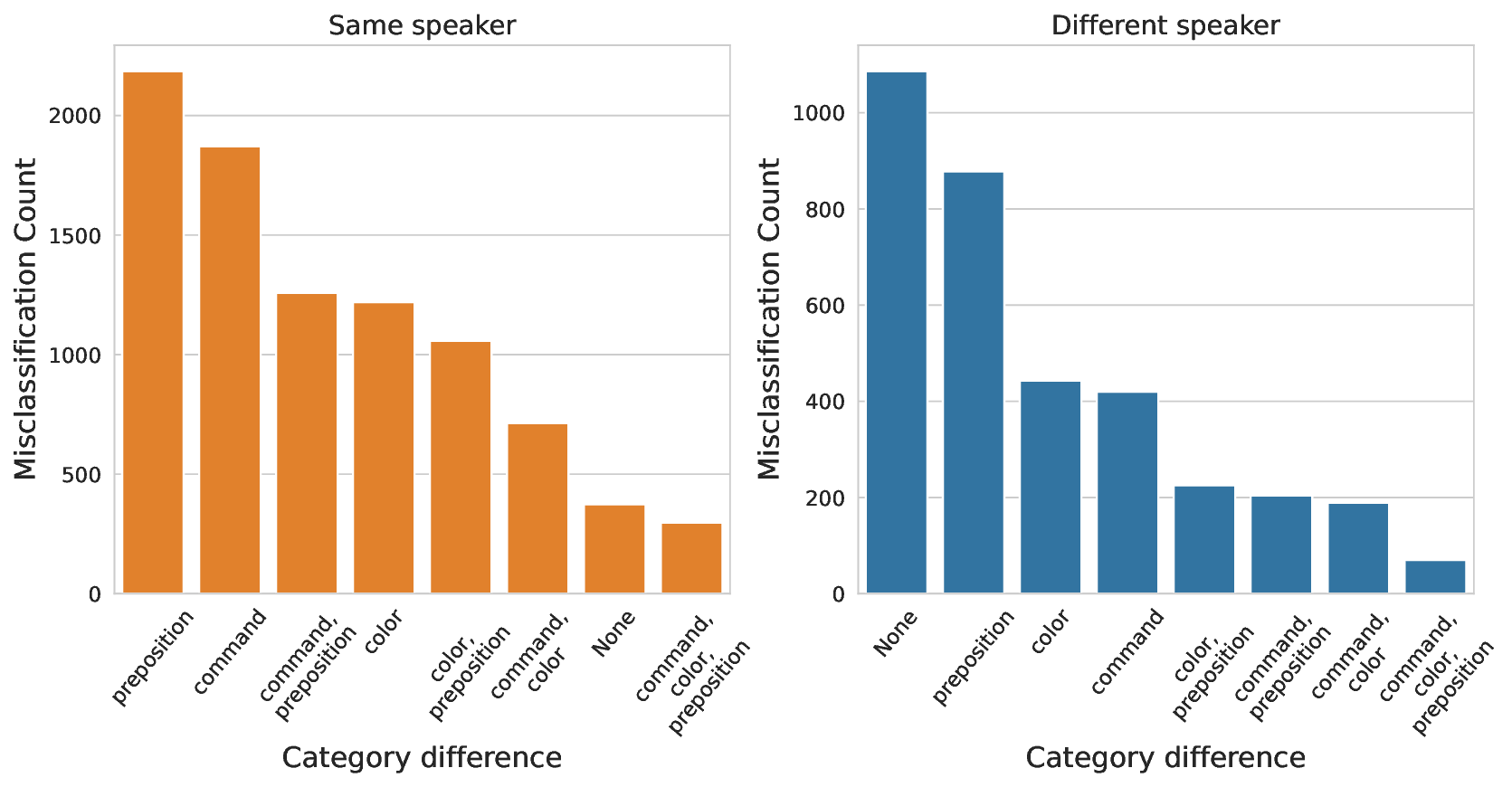}
	\caption{Most confused test set word categories.}
	\label{fig:most_confused_word_categories} 
\end{figure}

The proposed model's most confused phrases are analyzed by displaying the top 20 pairs of mistaken phrases for the same and different speakers, as shown in \autoref{fig:most_confused_phrases}. For the sake of simplicity, a single authentication phrase from the customized GRID dataset is denoted with a unique abbreviation which contains the first letter of the command word, color word, and preposition word. For example, the authentication phrase "place green by" from the modified GRID dataset can be denoted as "pgb". The majority of errors occur when the model confuses a single word from a word category, most notably the preposition category. It is worth noting that preposition category words are also the shortest ones included in our chosen sub-phrases. The count of errors made on pairs where the phrases differ by a word belonging to a specific word category is shown in \autoref{fig:most_confused_word_categories}. The x-axis displays the names of the word categories for which the pair phrases differ in word selection. None label is used when there is no difference between phrases. Fewer error are made for both the same and different speaker case as more words differ across phrases. This indicates that biometric security could be improved significantly by proposing strictly different authentication phrases across users. Most errors for the different speaker case are made when there is no difference in phrases as lip movements correspond to the same phrase and the model can't rely on it for discrimination.

\section{Conclusion}
\label{sec:conclusion}

This paper presents a one-shot deep learning approach for LBBA that, for the first time, models behavioral features both as style-of-speech and spoken phrase information. It improves over current one-shot approaches that model behavioral features only as style-of-speech which makes them vulnerable to video replay attacks of the client speaking any phrase. GRID dataset is used to obtain the right diversity and quantity of pairs needed to train a siamese neural network for our task. A siamese neural network based on 3D convolutions and recurrent neural network layers is trained for creating an input embedding compatible with a cosine similarity function. A custom triplet loss is used for efficient learning of the underlying representation. The model shows strong discriminatory power against both physical and behavioral features, achieving 3.2\% \textit{FAR} and 3.8\% \textit{FRR} on the test set of the custom GRID dataset. \textit{FAR} and \textit{FRR} are comparable to SOTA one-shot deep learning approaches relying only on style-of-speech, despite modeling a much more challenging problem. Analysis of physical and behavioral feature impact on performance showed that the model has the greatest confidence in its prediction when both physical and behavioral features are strictly the same or strictly different. Furthermore, results on pairs where either physical or behavioral features are different show that our model is better at discriminating based on behavioral features than physical ones. When distinguishing phrases, the biggest percentage of errors are made with phrase pairs that differ by a single word located at the same word index inside the phrase. Possible future work includes  exploring authentication phrase engineering at the viseme level and evaluating the approach in more challenging conditions.

\bibliographystyle{elsarticle-num} 
\bibliography{LBBA_bibliography}

\appendix
\newpage
\section{Distributions of test set pairs prediction scores}
\label{a:sec:miss}

\begin{figure}[htbp]
	\centering
	\includegraphics[width=\linewidth]{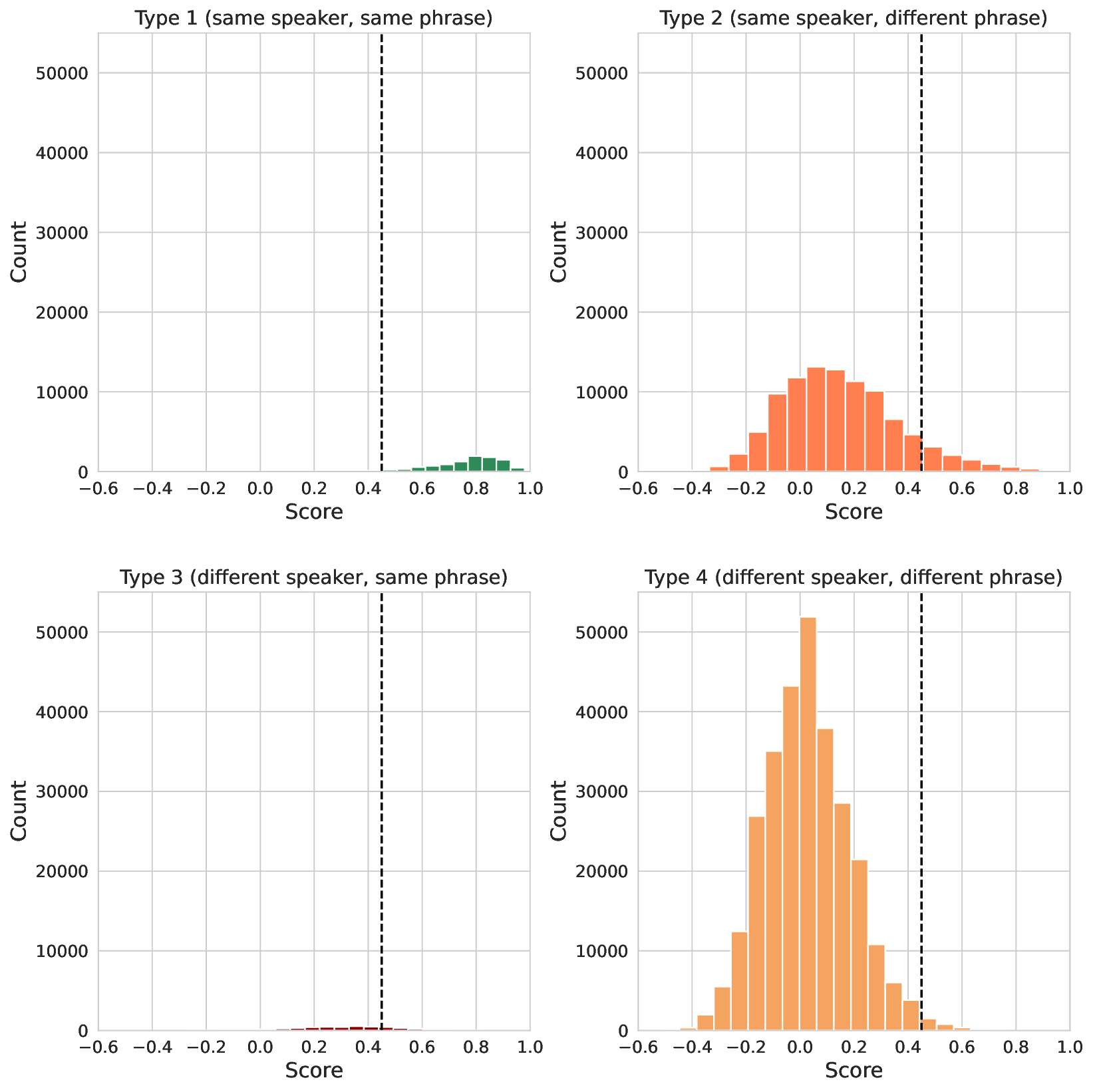}
	\caption{Histograms of test set pair prediction scores, with respect to pair type.}
	\label{fig:histogram_scores_separated} 
\end{figure}

\begin{figure}[htbp]
	\centering
	\includegraphics[width=\linewidth]{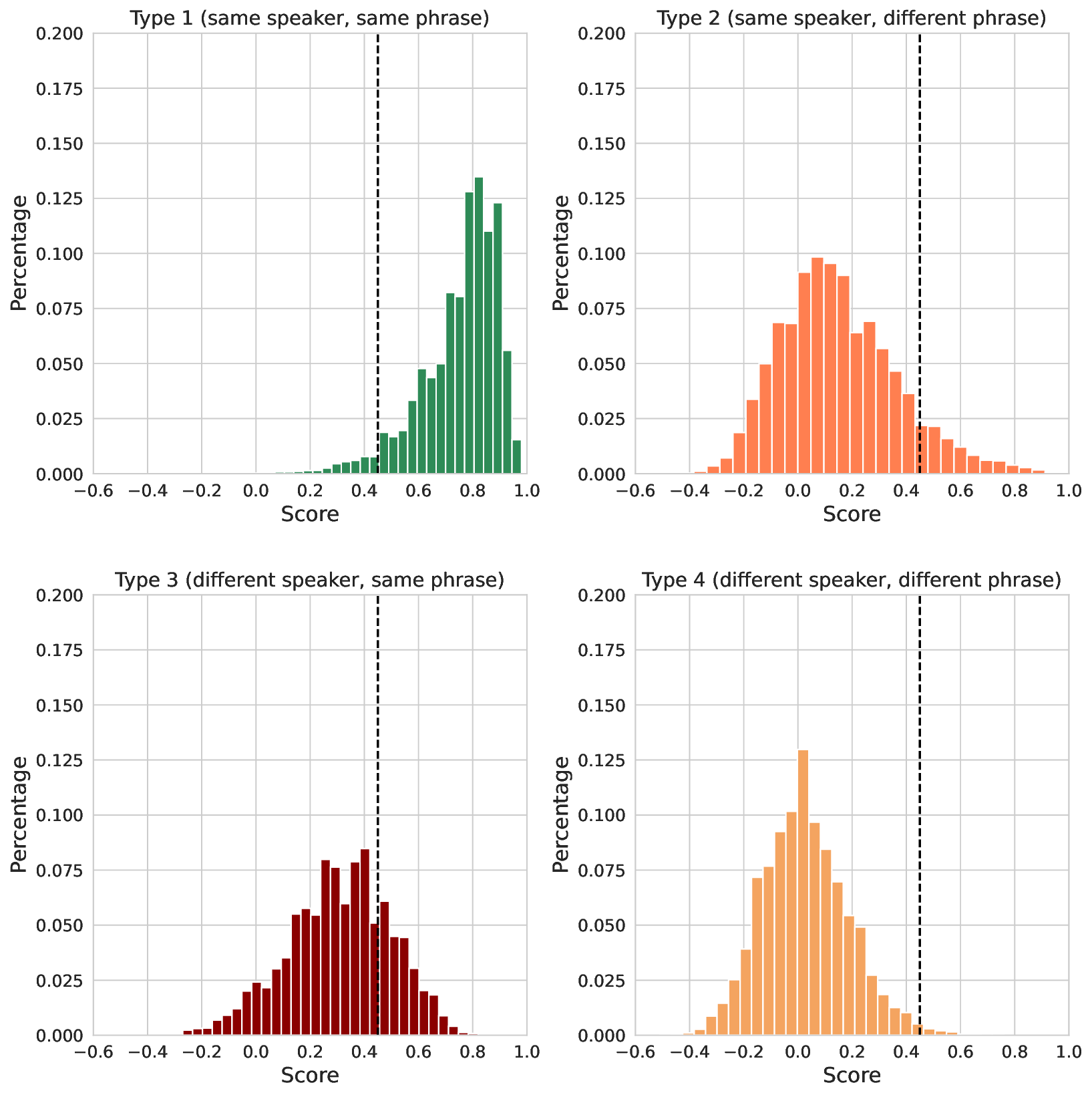}
	\caption{Normalized distributions of test set pair prediction scores, with respect to pair type.}
	\label{fig:histogram_scores_separated_normalized} 
\end{figure}

\end{document}